\documentclass{sig-alternate-2013}

\permission{\copyright 2017 International World Wide Web Conference Committee \\ (IW3C2), published under Creative Commons CC BY 4.0 License.}
\conferenceinfo{WWW 2017,}{April 3--7, 2017, Perth, Australia.}
\copyrightetc{ACM \the\acmcopyr}
\crdata{978-1-4503-4913-0/17/04. \\
http://dx.doi.org/10.1145/3038912.3052681  \\
\includegraphics{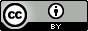}%
}
\usepackage{acmcopyright}
\usepackage{times}
\usepackage[hyphens]{url}
\usepackage[implicit=false]{hyperref}
\usepackage{latexsym}
\usepackage{amsmath}
\usepackage{graphicx}
\usepackage{color}
\usepackage{caption}
\usepackage[dvipsnames,table]{xcolor} %
\usepackage{xspace}
\usepackage{booktabs}
\usepackage{float}

\newcommand{\xhdr}[1]{{\noindent\bfseries #1.}}

\newcommand{\cut}[1]{}

\newcommand{\correctness}{correctness\xspace}

\newcommand{\sep}{\noalign{\vskip 0.8mm}}
\newcommand{\improvement}{synergy\xspace}
\newcommand{\Gteam}{G_{\mathrm{team}}}
\newcommand{\pp}[1]{\mathrm{p_{\mathrm{#1}}}}
\newcommand{\Gp}[1]{G_{\mathrm{{\pp{#1}}}}}
\newcommand{\dist}{\mathit{dist}}

\newcommand{\specialcell}[2][c]{%
  \begin{tabular}[#1]{@{}l@{}}#2\end{tabular}}

\title{When Confidence and Competence Collide: \\ Effects on Online Decision-Making Discussions}

\numberofauthors{3}
\author{
\alignauthor
Liye Fu \\
\affaddr{Dept. of Computer Science}\\
       \affaddr{Cornell University}\\
       \email{liye@cs.cornell.edu} \\
\alignauthor
Lillian Lee\\
       \affaddr{Dept. of Computer Science}\\
       \affaddr{Cornell University}\\
       \email{llee@cs.cornell.edu} \\
\alignauthor
\mbox{Cristian Danescu-Niculescu-Mizil}\thanks{Corresponding senior author.} \\
       \affaddr{\mbox{\hskip 0.35in Dept. of Information Science}}\\
       \affaddr{\mbox{\hskip 0.35in Cornell University}}\\
       \email{\mbox{\hskip 0.35in cristian@cs.cornell.edu}}\\
}

\begin{document}

\setcopyright{acmcopyright}

\maketitle

\begin{abstract}
Group discussions are a way for individuals to exchange ideas and arguments in order to reach better decisions than they
could on their own. One of the premises of productive discussions is that better solutions will prevail, and that the
idea selection process is mediated by the (relative) competence of the individuals involved. However, since people may
not know their actual competence on a new task, their behavior is influenced by their self-estimated competence --- that
is, their {\em confidence} --- which can be misaligned with their actual competence.

Our goal in this work is to understand the effects of confidence-competence misalignment on the dynamics and outcomes of
discussions. To this end, we design a large-scale natural setting, in the form of an online team-based geography game,
that allows us to disentangle confidence from competence and thus separate their effects.

We find that in task-oriented discussions, the more-confident individuals have a larger impact on the group's decisions
even when these individuals are at the same level of competence as their teammates. Furthermore, this unjustified role of
confidence in the decision-making process often leads teams to under-perform. We explore this phenomenon by
investigating
the effects of confidence on conversational dynamics.
For example, we take up the question: do more-confident people introduce more ideas than the less-confident, or do they
introduce the same number of ideas but their ideas get more uptake?
Moreover, we show that the  language people use is more predictive of a person's confidence level than  their actual
competence.  
This also suggests potential practical applications,
given that in many settings, true competence cannot be assessed
before the task is completed, whereas the conversation can be tracked during the course of the problem-solving process.
\end{abstract}


\vspace{0.5cm}

\section{Introduction}
\label{sec:intro}
\newcommand{\introcompetence}{competence\xspace}
\newcommand{\introcompetent}{competent\xspace}

It is hard to argue with Griffin and Tversky's \cite{Griffin:CognitivePsychology:1992} opening statement, ``The weighing
of evidence and the formation of belief are basic elements of human thought''.\footnote{This is not to say these are {\em all} the elements of human thought.}  
Griffin and Tversky go on to say that
one of the major findings from psychology, philosophy and statistics regarding how to evaluate evidence
and assess confidence is that ``people'', ranging from the ordinary to the expert, ``are often more confident in their
judgments than is warranted by the facts''.\footnote{See also, among others,   \cite
{Koriat:JournalOfExperimentalPsychologyHumanLearningAnd:1980,kruger1999unskilled}, the
latter giving rise to the phrase ``Dunning-Kruger effect''.}  At the same time, many studies have also
found underconfidence to be rampant, as well; for an entry into that literature, see Erev et al.~\cite
{erev1994simultaneous}, or, for an intriguingly titled study, \cite
{Lichtenstein:OrganizationalBehaviorAndHumanPerformance:1977}.  Thus, confidence and competence are often
misaligned.\footnote{``One of the painful things about
our time is that those who feel certainty are stupid, and those with any imagination and understanding are filled with
doubt and indecision.
I do not think this is necessary.'' --- Bertrand Russell}

What 
effect does this misalignment of confidence and competence have on
\begin{itemize}
\item[(a)] group  decision-making,  and  \item[(b)]
the dynamics of group discussions?
\end{itemize}
  With respect to (a), for example, who has more influence on the group's decision: those who are
overconfident
but under-\introcompetent, or those who are
\introcompetent but less confident?
Similarly,
with respect to (b):  do the overconfident but under-\introcompetent
tend to dominate the entire conversation, or do the more-\introcompetent eventually
have more
say?
Studies on (a) \cite
{Neale:AcademyOfManagementJournal:1985,Sniezek:YobhdCjaJidOrganizationalBehaviorAndHumanDecisionProcesses:1989,Zarnoth:JournalOfExperimentalSocialPsychology:1997}
have generally shown that the overconfident have greater clout and that overconfidence leads to sub-optimal
outcomes; but these experiments have been relatively small in scale or involve simulations.
Topic (b) appears understudied.

In contrast, {\em the work we present in this paper represents a first investigation into both the conversational and
decision-making impact
of
confidence-competence misalignment
at a large scale and in an online natural setting.}
Our experimental platform is an online
collaborative
geography
puzzle game that has been played by over 10,000 teams;
in the game, players have potentially different
``views''
of a given location and try to determine that location by pooling their information.
The system records each individual's guess before 
 they interact
 with other teammates, 
each individual's confidence in their guess,  
the entire team conversation,
and the accuracy of the final answer that the team settles on.  (See Section \ref{sec:data:measure}
for more details.)

In our data, more than half the users indicate a confidence level that does not match their \introcompetence level, with
these 
miscalibration
 errors being committed both by people with 
 low competence and by people with high competence.
(See Section \ref{sec:miscalibration} for more details.)
We thus have a substantial number of instances of under- and over-confidence to analyze in the context of group
 decision-making discussions.

Our findings with respect to point (a) --- effects on group decision-making --- are that: beyond
\introcompetence, confidence gives people additional control over team decisions (Section
\ref{sec:interaction:decisions}). While this result is in line with those of \cite
{Zarnoth:JournalOfExperimentalSocialPsychology:1997}, we additionally explicitly examine the impact of degrees of
confidence
differences between group members, controlling for correctness.
We see that
{\em when the most
\introcompetent
individual in a
team is less confident than the least \introcompetent one, the team generally performs worse compared to teams with the same
composition in terms of \correctness but in which individuals have accurate confidence estimates.} This suggests that
teams where misalignment exists between individuals' relative confidence and relative \introcompetence
fail to
reach their potential in terms of performance (Section \ref{sec:interaction:performance}).

We are thus naturally led to delve into the novel research topic~(b) --- the interplay between confidence and the dynamics of
group
discussions --- to help understand what happens in a group conversation that 
leads to
 the effects just described.
{\em For instance, we examine the role confidence plays in the idea-selection process: do more-confident people introduce more ideas than the less-confident, or do they
introduce the same number of ideas but their ideas get more uptake?}
We find that in fact, more-confident
individuals tend to introduce more ideas in discussions, even when there are not any more competent than their teammates
(Section~\ref{sec:interaction:mediate}).

Finally, as a practical contribution, we show that there are linguistic cues that are predictive of a
person's confidence level, suggesting
that interfaces for group decision-making could potentially make use of confidence information to help in the collaboration process
(Section \ref{sec:eval}). 

\smallskip
\noindent {\bf Terminological note: from ``competence'' to ``correctness''.} The term ``competence'' is used in the related
psychology
literature to indicate a person's actual 
observed performance
 in a given task, and we would have preferred to be consistent with that literature.  However, in this paper, we are
discussing both
``confidence'' and ``competence'', and the two words unfortunately (and perhaps ironically) sound so similar as to potentially introduce
confusion.
 We therefore use the term ``correctness'' instead of ``competence'' in most of the remainder of this paper.

\section{Experimental Setup}
\label{sec:data}

\subsection{Large-scale online discussion setting}
\label{sec:data:streetcrowd}

To study the role 
confidence plays in the dynamics of online discussions and decision-making 
processes,
  we need a natural setting where we can gather participants' confidence labels and observe
  details of the group decision-making process.
With these constraints in mind, we design an experimental platform in the guise of an online team-based world exploration game, StreetCrowd \cite{niculae16constructive}.\footnote{This game has been approved by the Cornell University's IRB 
(Protocol 1504005555)
and is currently live at {\scriptsize \url{http://streetcrowd.us/start}}; an unaffiliated gameplay video is available at {\scriptsize \url{https://youtu.be/Lp47w-lWsn0}}.}

StreetCrowd is played in teams of at least two players, and is built around a geographic puzzle:
the system chooses some spot on Earth, and players are given access (only) to ground-level Google
StreetView images of that spot, which they can navigate to try to determine where the spot is.
Each game
consists of two stages:

\begin{itemize}
\item {\bf Solo phase -- exploration but no discussion. }
There is no communication between players at this stage.
 Each player has three minutes to  independently explore the surrounding environment and find information that may
help them locate the spot,
e.g., geographical landmarks, apparent climate, whether cars drive on the left-hand or the right-hand side of the road,
and so on.
The player
makes an {\em individual guess} by placing a marker on an interactive world map. To proceed to the team phase, the
player is prompted for a reason
(in a few words
and 
a confidence level for her guess. 

\item{\bf Team phase -- discussion but no exploration. } At this point, further navigation is disabled,
but players in a team are
 placed in a chat room where they exchange ideas and opinions in an attempt to decide on a single {\em team guess}. They have access to a shared map and to a marker that each one of them can move.  When all players have agreed on the position of the marker on the map, or the time limit is reached, the game ends.
(See Table~\ref{tab:chat_example} for an example team discussion.)
\end{itemize}

\begin{table}[!t]
\centering
  \caption{Excerpt from a chat during the team phase. The true answer for this puzzle is France. Player 1 has guessed the country correctly, but is only sure the location is in Europe. Player 2 and 3 guessed New Zealand and Nepal respectively. During the chat, players attempt to convince each other, while making intermediate team marker placements during the process. In this example, player 1 tried to persuade the team to guess France whereas the other two players argued for New Zealand.}
  \label{tab:chat_example} 
  \begin{tabular}{ l l }
  \toprule
  {\it player 1}:& its in the alps somewheres \\ \sep
  {\it player 2}:& Crazy amount of sheep \\ \sep
  {\it player 2}:& I put new zealand \\ \sep
  {\it player 3}:& my sources say no \\ \sep
  {\it player 1}:& well maintained mnt road says europe \\ \sep
  {\it player 2}:& good point \\ \sep
  {\it player 3}:& new zealand is good shout \\ \sep
  {\it player 2}:& the land looks a bit arid though \\ \sep
  {\it player 2}:& thats why i thought nz \\ \sep
  {\it player 1}:& well NZ is like europe in that way \\ \sep
      & but dont notice any of their typical vegitation \\ \sep
  {\it player 3}:& nepal \\ \sep
  {\it player 2}:& I think the road is too nice to be in nepal \\ \sep
  {\it player 1} & {\it has moved the team marker [to France]} \\ \sep
  & $\cdots \cdots$  \\ \sep 
  {\it player 2}:& and yeah, nz is like europe but not in europe lol \\ \sep
  {\it player 2}:& lol \\ \sep
  {\it player 2} & {\it has moved the team marker [to New Zealand]} \\ \sep
  & $\cdots \cdots$  \\ \sep 
  \bottomrule
  \end{tabular}
\end{table}

In this setting, we can generate a virtually infinite number of puzzles with known correct answer --- the location $L_{
{\rm true}}$ used to generate the respective StreetView.  We can also directly measure how correct a guess $G$ is based
on the
 the distance to the true location
  $\mathit{dist}(G, L_{{\rm true}})$ 
  (we use great-circle distance). %
 We also record a discrete level of correctness based on how precise the guess is: we use reverse geocoding to compare
 the coarse geographical location of the guess 
 with that of the correct answer
 (e.g., do they both fall in the same country?),
 and categorize guesses into the four levels of 
 \correctness,
 which we call {\em guess precision}, depicted in
 Table~\ref{tab:correct_level}.

Over the past year, StreetCrowd has accumulated nearly
12,600
 team games (comprising about 30,000 individual-phase games) involving more than
 5,800
  unique players.
   We discard games in which individual players fail to make a guess,
   the team does not agree on a final guess,
    or no communication occurs. We also
    heuristically
    filter out games that may involve cheating and games 
    for
    which we could not identify the coarse location of a player's guess
    automatically.
\begin{table}[!t]
\centering
  \caption{Numerical correctness levels based on guess precision.}
  \label{tab:correct_level} 
  \begin{tabular}{ l  c }
  \toprule
  \textbf{guess precision} & \textbf{\correctness level}  \\
  \midrule
  wrong continent & 1 \\
  correct continent, wrong country & 2 \\
  correct country, wrong region& 3\\
  correct region & 4 \\
  \bottomrule
  \end{tabular}
\end{table}

\begin{table}[!t]
\centering
  \caption{Possible choices for self-estimating guess precision and corresponding numerical confidence levels.  For
  validation, we also show average zoom levels used by  players declaring the respective confidence levels.
  (At zoom level 1, the entire world is displayed,  whereas at zoom level 10, players are looking
at roughly the city level.)}
  \label{tab:conf_level}
  \begin{tabular}{ l  c  c}
    \toprule
    \textbf{estimated precision} & \textbf{confidence level} & \textbf{zoom level}\\ \midrule
    ``Could be anywhere'' & 1 & 4.3 \\
    ``I know the continent!'' & 2 & 4.6 \\
    ``I know the country!!'' & 3 & 5.4 \\
    ``I know the region!!!'' & 4 & 7.8 \\
    \bottomrule
  \end{tabular}
\end{table}

\subsection{Measuring confidence}
\label{sec:data:measure}

Studies of confidence in the  psychology literature have largely considered three possible operationalizations of
confidence \cite{moore2008trouble}:

\begin{enumerate}

\item Confidence as a self-estimation of (a concrete measure of) performance, that is, how well one did or will do.
Examples include
a student's estimate of 
her score on an exam~\cite{Boud:HigherEducation:1989,clayson2005performance},
 or an estimate of the time it takes to finish a particular task~\cite{buehler1994exploring}.

\item Confidence as the level of certainty in one's answer. This type of confidence can be extracted by, for example,
asking participants to indicate how sure
they are of their answer on an ordinal scale \cite{Moussaid:PlosOne:2013}.

\item Confidence as placement of one's ability relative to others. Studies that use this definition of confidence may ask participants to estimate
their performance percentile with respect to all the other participants
\cite{larrick2007social}, e.g., ``top 5\%''.

\end{enumerate}

\xhdr{Employing operationalization 1} In our work, we
focus on the first operationalization of confidence, because it concerns a direct comparison between an
individual's actual performance and 
her 
estimate of 
her
performance.  Misalignment between the two can  then be
interpreted
in terms of {\em overconfidence} --- when the individual's estimated performance is above 
her
actual performance ---
and {\em underconfidence} --- when the estimated performance is under the actual performance. For instance, player 1 in Table~\ref{tab:chat_example} exemplifies underconfidence as she has guessed the correct country, yet she only estimated her correctness to be at the continent level.
To capture this notion of (over)confidence in StreetCrowd, we compare {guess precision} with {\em estimated precision},
the four levels of which are given in Table~\ref{tab:conf_level}. At the end of the solo phase, players are prompted to
indicate
how
good 
they think their guess is by clicking  on one of the four estimated-precision buttons,  whose labels are shown in the
leftmost column of Table~\ref{tab:conf_level}.
The correspondence between the correctness levels in Table~\ref{tab:correct_level} and confidence levels in Table~\ref{tab:conf_level} is deliberate, because \emph{the signed difference between the numerical confidence level and the numerical
correctness level gives us a  degree of misalignment}: a larger confidence level relative to correctness
level (e.g., ``I know the country'' but the guess was really ``correct continent, wrong country'') indicates overconfidence;
when the confidence level is smaller than the correctness
level, we have underconfidence;
 the two values being equal indicates 
correct~calibration.

To check if people are indicating their self-evaluated confidence reasonably --- as opposed to clicking randomly or
primarily based on how the buttons are placed in the interface --- we compare the  guess-precision levels with  how
much the users zoomed in on the map before marking their guess.  Higher zoom levels reflect a greater level of
detail.   We expect players who believe their guess is more precise to zoom more, and indeed in our
data we do observe that players with higher confidence levels tend to have higher zoom levels on average (Spearman's
$\rho$ = 0.30, $p$-value < 0.001, Table \ref{tab:conf_level}).\footnote{We use Spearman's $\rho$
  throughout our study for reporting correlations. Kendall's $\tau$ 
  gives similar qualitative result.}

\xhdr{The other operationalizations}
To compare with relevant literature,
in an earlier version of StreetCrowd,
we also evaluated
confidence according to the second operationalization listed above: how certain an individual is in 
her
answer. We
collected these labels from about 1,800
games by prompting the individual to indicate 
the level of 
certainty in 
her 
guess on a scale from 0\% to 100\% using a slider.
(We excluded users who did not move the slider at all.)
However, we abandoned this design in favor of the estimated-precision approach outlined above: offering users the
relatively objective comparison points of continents, countries, and within-country regions\footnote{We take states returned by Google and Bing geocoding services as regions.}  
was preferable to trying to
interpret what, say, ``65\%'' means or deciding whether one person's ``65\%'' could definitely be considered more
confident than another person's ``60\%''.

In this work we do not consider the third operationalization of confidence --- 
relative placement among others --- which
requires the individuals to jointly consider other people's correctness as well as their own, 
for two reasons: 
(1)
this would require post-hoc surveys of the participants after playing the game, which would be harder to collect in a
natural setting,
and (2) it would be hard to distinguish whether misalignments between estimated and
actual performance are
caused by  participants  misjudging their own ability, or misjudging the ability of the others.

\section{Confidence miscalibration}
\label{sec:miscalibration}
As mentioned in the Introduction, a person's confidence does not necessarily align with 
her 
actual objective ability.  Earlier studies suggest that this miscalibration follows systematic patterns driven by the level of competence of the individuals.
For instance, several researchers \cite{kruger1999unskilled,Moussaid:PlosOne:2013} argue that one needs a certain level of competence in order to reasonably calibrate one's confidence, and that as a result, non-competent individuals tend to overestimate their performance.
Indeed, an extensive series of studies from the field of education dating back to the 1930's \cite{sumner1932marks} 
 (see \cite{Boud:HigherEducation:1989} for a comprehensive survey) 
consistently indicate that when estimating their grades, ``weak'' students tend to overestimate their grades, whereas ``strong'' students have a (less pronounced) tendency to underestimate their grades.
In this section, we use our online setting to analyze the nature of this systematic bias in confidence miscalibration in more detail, and at a much larger scale than was previously possible.  This builds the backdrop
against
which we later examine its consequences on the dynamics and outcomes of decision-making discussions.

We examine misalignments with respect to two different common operationalizations of confidence introduced earlier in Section~\ref{sec:data:measure}: 

\xhdr{Certainty in one's answer}
For the subset of the data where players indicated their confidence
via a slider, we group players by their \correctness level (as listed in Table \ref{tab:correct_level}).  For each such \correctness group,
  we compute the correlation between 
the actual distances between 
  each participant's initial guess
  and the true location of the puzzle, $\dist(G, L_{{\rm true}})$, and their indicated level of confidence.  We find that for players whose location was on the wrong continent, their self-reported confidence level has no significant correlation with distance between their guess and the true location (Spearman's $\rho$ = -0.006, $p$-value = 0.88),
  whereas in contrast, the correlation is significant for all other better-performing groups. In fact, the correlation is the strongest for the group of players who hit the correct region ($\rho$ = 0.41, $p$ $<$ 0.001). This confirms that one needs a certain level of competence to reasonably evaluate one's performance. 

\xhdr{Self-estimation of performance}
Since the previous operationalization of confidence-as-certainty is not rooted in an objectively measurable outcome, misalignments cannot be directly translated into under- or over-confidence.  For this reason, for the remainder of this paper, we  operationalize confidence as a self-estimation of performance, a common interpretation in the psychology and education literature \cite{Boud:HigherEducation:1989,kruger1999unskilled,moore2008trouble}.
This allows us to non-ambiguously map misalignment between the estimated performance and the actual performance directly to under- and over-confidence.

\begin{figure}[!t]
\centering
\includegraphics[scale = 0.39]{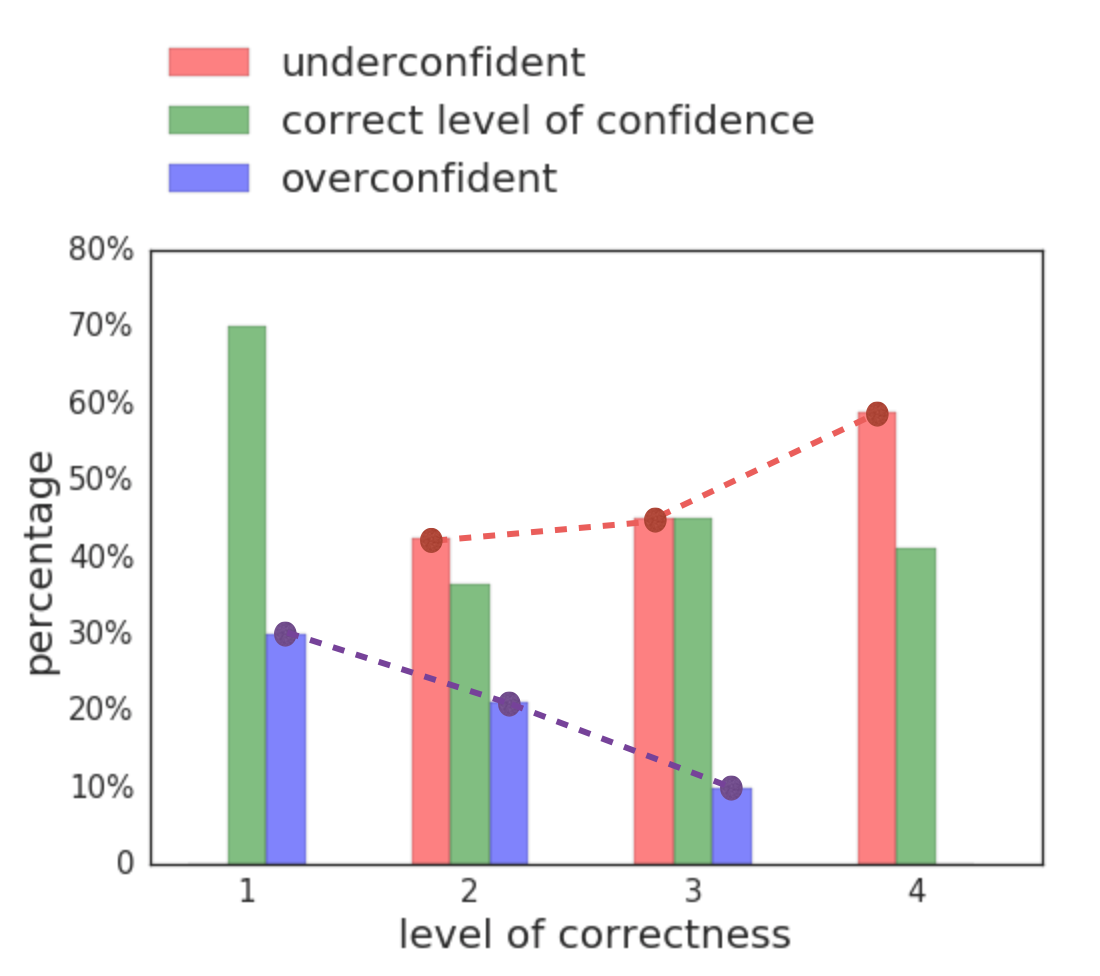}
\caption{
Misalignment between competence and confidence. For each correctness level: the fraction of players who are underconfident, at right level of confidence, and overconfident.  More competent individuals are less likely to be overconfident.
}
\label{fraction}
\end{figure}

More than half of the users fail to correctly estimate their performance.
Figure \ref{fraction} breaks this failure down by
correctness level,
and shows that indeed, in agreement with  \cite{kruger1999unskilled,Moussaid:PlosOne:2013},
individuals that are more wrong are more likely to overestimate their performance (blue dashed line),
whereas increasing correctness corresponds to increasing underconfidence.

We note here two experimental limitations that are intrinsic to any experimental study of confidence misalignment.  First, there is a boundary effect in that individuals achieving maximum performance on a task cannot be overconfident: e.g., a student that receives an exam grade of  A+ cannot produce an estimate of a higher grade; symmetrically, individuals that achieve minimum performance can not be underconfident.  Second, some individuals can appear to be under-confident simply by chance: e.g., a student answering a multiple choice question can select the correct answer by chance and simply appear to be more skillful than they really are. This effect does not apply
symmetrically
to over-confidence: a student knowing the answers to a multiple choice question is much less likely to select the incorrect answer by chance and appear less skillful than in reality.  This
asymmetry
alone could explain why underconfidence can appear to be more common than over-confidence.  Acknowledging that such limitations naturally apply to our setting as well, we refrain from making any claims regarding the relative propensity of under- and over-confidence from the data depicted in Figure \ref{fraction}.

\section{Confidence in team interaction}
\label{sec:interaction}

The promise of group over individual
deliberation lies in the intuitive notion that the more people we bring together, the more ideas for solving a
problem will be generated, and the more ideas generated, the more likely one
of them will provide an excellent solution \cite{Jackson+Poole:2003,laughlin2006groups,Valacich01021992}. Ideally, this process could lead to a collective 
performance
 that surpasses what individuals can achieve on their own, i.e., the interaction will have a {\em synergistic} effect.\footnote{
 See also the related concepts of collective intelligence \cite{woolley2010evidence} and assembly bonus effect \cite{collins1964social}.
  }.

In practice, however, this ``sum greater than the parts'' promise 
 is far from guaranteed \cite{niculae16constructive}.  The process through which solutions surface is complex and relies on constant negotiation between the participants, 
and ``process losses'' \cite{steiner1972group} can also occur. 
  While in an ideal setting the more competent individuals would have a larger say, in many realistic scenarios the actual level of competence on a new task is unknown.  The decision-making process is then driven in part by self-estimations of competence (which, as we have discussed above, can be heavily misaligned with actual competence),  
and decisions made through heuristics relying on such estimates can be worse than desired \cite{bang2014does}.
In this section we examine what are the consequences
of confidence (mis)alignment on the decision-making process, on the degree of resulting 
synergy
 and on the overall dynamics of the discussion.

\subsection{Confidence and relative influence}
\label{sec:interaction:decisions}

When a team makes a decision, different individuals can contribute to 
the 
proposed solution to different degrees. Here we examine the role confidence plays in regulating 
the relative influence 
different team members 
have on the final decision.

\xhdr{Relative influence} In our setting, we can estimate the influence on the team decision
by considering how much closer to 
her
individual guess the 
 player has managed to get the team guess
 to be, 
 relative to the other player.  More formally, we define influence of a given player $\pp{1}$ (with answer $\Gp{1}$)
 on a team answer $\Gteam$ relative to $\pp{2}$ (with answer $\Gp{2}$) by:

\[
\mathit{inf}(\pp1, \pp2) = \displaystyle\frac{\dist(\Gp{2}, \Gteam)}{\dist(\Gp{2}, \Gteam) + \dist(\Gp{1}, \Gteam)},
\]

\noindent noting that this measure of {\em relative influence} is agnostic to how correct the solution is.

For each team, we pair the player who achieved the highest level of correctness with the player that has the lowest level.
 Importantly, in cases in which all players have the same level of correctness, best and worst players are arbitrarily chosen.  For each team, we compute how much more control the best player has over the team decision relative to the worse player, i.e., 
 $\mathit{inf}(\pp{best}, \pp{worst})$.\footnote{For 
 clarity 
 we focus our
analysis
on the best and worst players in the team.  The majority 
($> 60\%$) 
of the games only have two players.}

\begin{figure}[!t]
\centering
\includegraphics[scale=0.34]{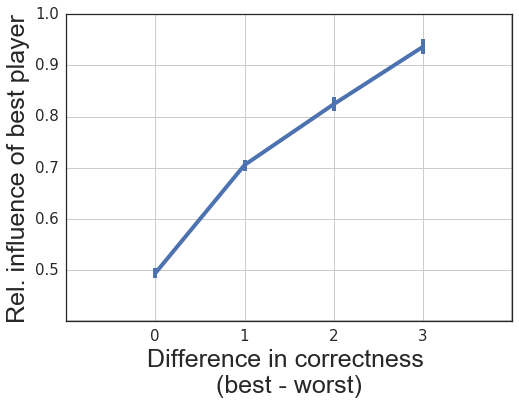}
\caption{The larger the difference in \correctness between the best and the worst player, the more relative influence the best player has on the final decision. A difference in \correctness of 0 corresponds to the case when all team members were at the same level of \correctness; in this case the ``best''  and the ``worst'' player have equal contributions (i.e., relative influence of 0.5), as expected since their roles are randomly assigned.  Throughout, error bars indicate standard error.
}
\label{mean_corr}
\end{figure}

\xhdr{Controlling for \correctness} As previously discussed, in order for the team discussions to be productive,
it is desirable to have the more competent member take more control and guide the team in their decisions.  It is comforting to see that, indeed, the best 
 player has a greater relative influence on the team solution than the worse player in the team: Figure \ref{mean_corr} shows that the larger the difference in \correctness between the best player and worst player, the more relative influence the best player has on the final decision.

Given that confidence and correctness are, at least to some extent, correlated ($\rho$ = 0.46, $p $-value $<$ 0.001),
it is important to control for the difference in \correctness in order to disentangle the effect of confidence.
To this end, we group games based 
on
the four possible values of the difference in \correctness and we compare the effect of confidence within these correctness-controlled groups.

\begin{figure}[!t]
\centering
\includegraphics[scale=0.35]{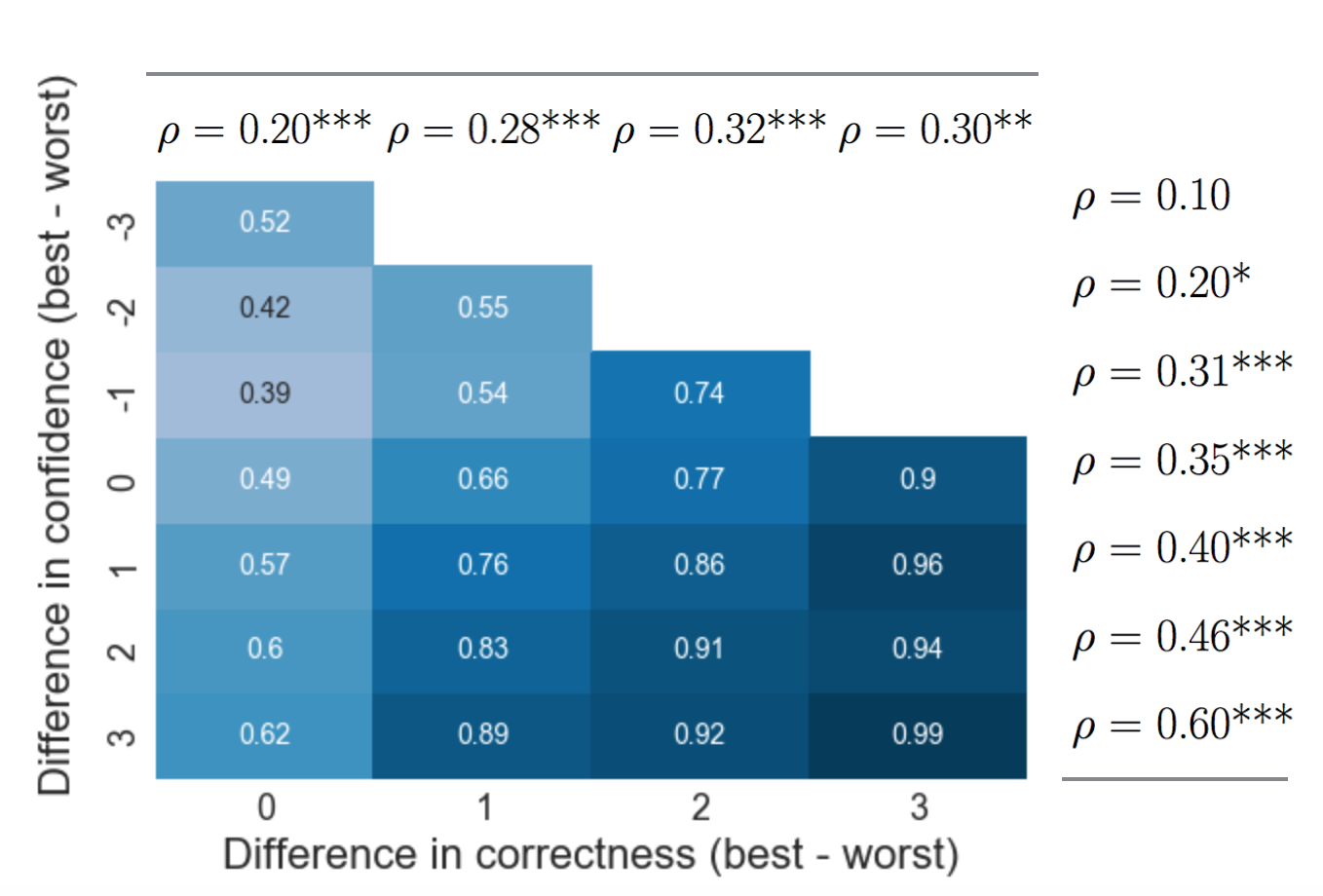}
\caption{The more confident the best player is relative to the worst player, the more influence (darker color) she has on the decision, even after controlling for difference in correctness (comparing across columns).  Spearman's correlation coefficients 
are computed using all data points within a group and are indicated, together with corresponding significance levels, for each column and row. Throughout, buckets with less than 15 instances are discarded. Statistical significance levels are indicated as: *:~$p<$ 0.05, **:~$p <$ 0.01, ***:~$p <$ 0.001.} 

\label{fig:rel_inf}
\end{figure}

\xhdr{Effects on relative influence} By comparing across correctness-controlled groups (columns) in Figure \ref{fig:rel_inf}, 
we observe that on average the more confident the best player is relative to the worst player, the greater  her influence on the final decision. Notably, this trend also holds even when the team members are equally correct 
(see 
first column, where 
``best'' and 
``worst'' roles are randomly assigned; $\rho$ = 0.21, $p$-value $<$ 0.001). {\em Beyond competence, confidence gives people unjustifiably greater control in the decision-making process.}

\subsection{Confidence and team performance}
\label{sec:interaction:performance}

When combined with the observation that non-competent individuals are likely to overestimate their competence (Section \ref{sec:miscalibration}), the result we just discussed suggests that misalignment between the competence and confidence of team members can prevent teams from achieving their 
potential. Driven by overconfidence, the least competent individuals can take control of the decision process, with harmful effects on the team outcome.

\xhdr{Measuring team \improvement}
While many
studies
 use correctness of the team answer as an indication of team performance
~\cite{coetzee2015structuring,friedberg2012lexical}, this type of measure neglects differences in initial team composition:
when a team of A students is doing better than a team of C students, it does not automatically mean they have collaborated better.
Here we focus instead on measuring how much a team improves over the potential of its members.%

To measure 
such improvement, or
{ \improvement},
for a team of $n$ players,
we first compare the distance between the team guess $G_{{\rm team}}$ and the true answer $L_{{\rm true}}$ to the average of distances between 
each player's 
individual guess to the true location. This gives the team a constructiveness score \cite{niculae16constructive}, which provides a control for the initial individual performance of the players in the team:
\[
c_{\rm avg} = \frac{1}{n}\sum_{i=1}^{n}{\dist(G_{\pp{i}}, L_{{\rm true}})} - \dist(G_{{\rm team}}, L_{{\rm true}}).
\]

In our setting, however, a team 
whose average guess is 
already close to the true location will have smaller maximum possible $c_{\rm avg}$ compared to teams that are farther from the correct answer.
This can happen either when a puzzle is relatively easy or when a team is composed of particularly strong players.
To account for this, we normalize the average score of the individuals in the team
to obtain a team {\em synergy} score: 

\[
s_{\rm avg} = \displaystyle\frac{c_{\rm avg}}{\frac{1}{n}\sum_{i=1}^{n}{\dist(G_{\pp{i}}, L_{\rm true})}}.
\]
Higher values indicate greater team 
improvement
over the individual performance of the team members.  We note that this measure can take negative values if the team's performance is worse than the average performance of its members.

\xhdr{Effects of confidence misalignment}
To test our intuition that 
misalignment between competence and confidence 
can be harmful to the decision-making process, we first consider 
the 227
teams where the worst player is 
overconfident 
to the point that she is more confident than the best player.
We find that 37\%
of these teams have negative 
synergy,  
compared to only 26\% of the rest of the teams (2-proportion z-test $p$-value < 0.001). 

\begin{figure}[!t]
\centering
\includegraphics[scale=0.35]{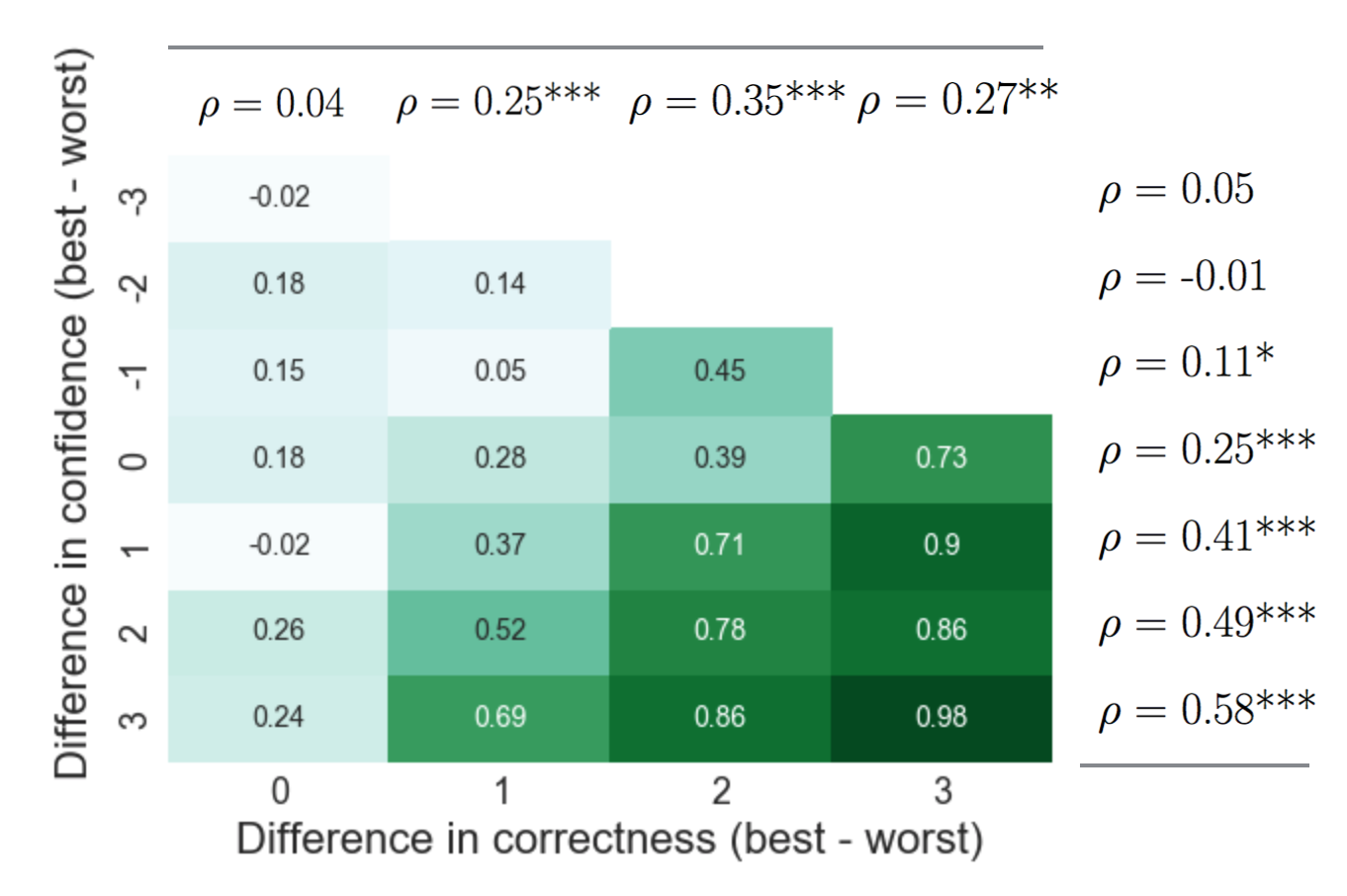}
\caption{When there is a real distinction between the best and the worst player (difference in correctness > 0), the more confident the best player is relative to the worst player, the greater synergy the team achieves (darker color), controlling for difference in correctness (comparing across columns).}
\label{fig:per_improvement}
\end{figure}

To explore the effect of misalignment more systematically, we compare the difference in confidence level of the best and the worst player, after controlling for correctness (Figure \ref{fig:per_improvement}).
Since we defined
synergy
as a symmetric team-based measure, we do not expect to observe any effect of confidence when the team members are equally correct (first column).
In all other cases, we notice that \emph{the larger the difference in confidence between the best and the worst player, the more productive the team discussion is: the player with the best solution is likely to lead the team towards 
better results}.  
In particular, teams where the confidence and correctness are misaligned such that the difference in correctness is larger than the difference in confidence --- i.e., the cells above the main diagonal (where difference in confidence equals difference in correctness) --- exhibit on average 
lower 
average \improvement.  This suggests that \emph{misalignment in confidence can be harmful to group discussions}, preventing groups from reaching their potential in terms of performance.

\subsection{How confidence mediates team interactions}
\label{sec:interaction:mediate}

The effects of confidence on team performance are intriguing, considering that we control for the actual competence of the individuals.  We now turn to examine the mechanisms through which confidence mediates group interactions, focusing on a 
 dimension of conversational dynamics that is particularly pertinent in the decision-making process: the introduction and discussion of ideas~\cite{niculae16constructive,Zhang:Naacl:2016}.

We consider two possible hypotheses that could explain the observed effects of confidence-competence alignment:
\begin{enumerate}
\item[\textbf{Hyp 1}] More-confident individuals contribute more ideas.  If the best individual is also the most confident one, the team benefits from having better quality ideas to consider in order to make steady progress to better solutions.
\item[\textbf{Hyp 2}] Ideas introduced by less-confident individuals are easier to discard.  
If competence and confidence are aligned, 
this can help the team focus on the ideas introduced by the more 
competent 
individuals.
\end{enumerate}

To gather valid ideas for a particular puzzle, we consider all nouns, proper nouns and adjectives that are not stopwords as candidate ideas. 
If any of these candidate ideas has been adopted (i.e., mentioned by at least two different participants) by any team attempting the respective puzzle it is included in the set of ideas for that puzzle. 
Example ideas discussed in the game illustrated in Table \ref{tab:chat_example} include ``sheep'', ``alps'' and ``arid''.
We also add to this puzzle-specific collection a set of geographical terms,\footnote{Source: Wikipedia's glossary of geography terms.} such as ``Nepal'', to provide better coverage of potential ideas.
 While we find this simple method effective for our particular setting, we acknowledge that more sophisticated methods for detecting, representing and tracking ideas are an important direction for future work (Section \ref{sec:discussion}).

\begin{figure}[!t]
\centering
\includegraphics[scale=0.38]{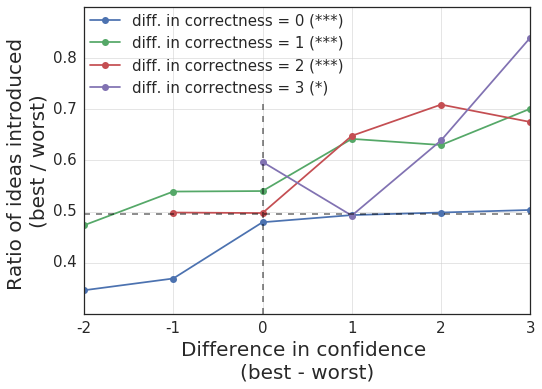}
\caption{
The more confident the best player is, the more ideas she introduces 
(relative to the worst player). Horizontal  dashed line indicates equal contribution of ideas by the best and worst player. Vertical dashed line indicates the case when best and worst individuals are equally confident. 
Significance level for Spearman's correlation coefficient is indicated for each 
difference in correctness level (each line). 
}
\label{fig:per_idea}
\end{figure}

To investigate the effect of relative confidence on the number of new ideas a player introduces to a discussion, we compare the ratio between the number of ideas introduced by the best player and the number of ideas introduced by the worst player, controlling for difference in correctness. 
We find that the more confident the best player is relative to 
her teammate
(larger difference in confidence), the more ideas 
she tend to 
introduce
relative to their teammates (Figure \ref{fig:per_idea}).  Particularly noticeable is the case where there is no difference in correctness between team players (difference in correctness = 0, $p$-value $<$ 0.001):  the more confident, but equally correct, individual 
still introduces
more ideas. 

It may be conceivable that the difference in number of ideas introduced may simply be explained by confident players talking more during the games. Yet, there is no clear correlations between number of words spoken and player confidence, indicating that it is the ideas in the chat messages that are more important. 

\begin{figure}[!t]
\centering
\includegraphics[scale=0.38]{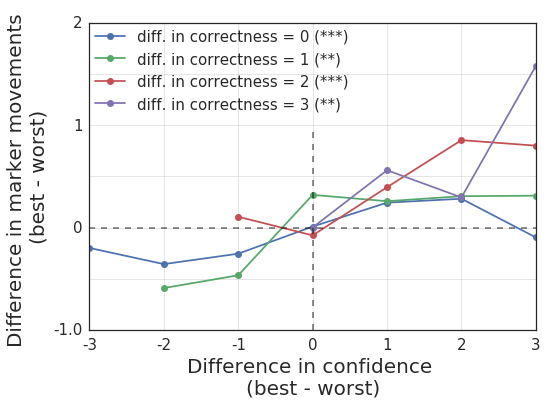}
\caption{The more confident the best player is, the more marker movements she makes relative to the worst player. Horizontal dashed line indicates equal marker placements.}
\label{fig:diff_marker}
\end{figure}

Since our setting also allows players to suggest solutions by moving the team marker during the discussion, we also examine marker movements as an alternative means of understanding the dynamics of the decision-making process that is less sensitive to the noise inherent to extracting ideas from natural language.
 We  observe similar trends as in the case of idea introduction: the more confident the best player is relative to her worst 
 teammate, 
 the more she contributes to team's marker movements
  (Figure \ref{fig:diff_marker}).  

Together, these results support the first hypothesis linking confidence with conversational dynamics: confidence seems to regulate the exchange of ideas in decision-making discussions, such that the more-confident individuals introduce more ideas.

The effect of confidence in filtering and selecting ideas in teams is less clear. 
We find no significant correlation between the level of confidence and the percentage of ideas that are eventually adopted by other team members.

\section{Linguistic Cues of Confidence}
\label{sec:eval}
So far we have discussed how confidence mediates team interaction and performance.  However, in online chats, the only means that a person (or a system) has to judge or estimate  the confidence of another is by picking up cues from what the other person says.
Therefore, in this section we ask
 whether there are linguistic cues that are 
indicative
 of an individual's confidence.  
 Motivated by our previous results, we focus on identifying signals that can not simply be explained by actual competence.
%

%

%
%

%
%
%
%
%
%
%
%
%
%
%
%
%
%
%
%
%

%
%
%
%
%
%
%
%
%
%
%
%
%
%

%
%

%

%
%
%
%
%
%

%

%
%
%
   
%

%
   
%

%

%

%
   
%

%
 
%

%

%
%

%

%
    
%
     
%
%
%
    
%
%
    
%
    
%
%
%
%
%
%
%

\begin{table*}[!t]
\rowcolors{1}{white}{lightgray}
\centering
\caption{Linguistic cues for confidence, placed in post-hoc categories.   The ``more-confident'' and
``less-confident'' columns list words that are among the top 30 features in at least half the test folds; daggers
mark words that are among the top 10 features in at least half the test folds. In the example utterances, word of the type represented by the row are shown in bold.}

  \begin{tabular}{ l p{3cm} p{3cm} p{3.5cm} p{3.5cm}}
    \toprule
    \textbf{category} & \textbf{more-confident} & \textbf{less-confident} &   \textbf{more-confident example} & \textbf{less-confident example}\\ \midrule
   
    personal pron. & we &  me, you$^{\dagger}$ & \specialcell{i think {\bf we} just go up \\ in the mountains} & looks like thailand to {\bf me}. \\ \sep
  
    interrogatives & which & \specialcell{how, what$^{\dagger}$, \\ where, why} & \specialcell{i found some writing {\bf which} \\ looked mandarin} & {\bf what} do you think ? \\ \sep

    thinking (tense) &
   \specialcell{think} & 
   
   \specialcell{thought, thinking$^{\dagger}$} & i {\bf think} it is nearer to the ocean &  

   \specialcell{i was {\bf thinking} brazil too} \\  \midrule

    hedges and modals & \specialcell{probably, maybe, \\ might, should, can} &  guess, could & 

    it {\bf might} be too far north & lol i {\bf guess} so \\ \sep
    
    certainty & \specialcell{100[\%], actually,\\ pretty [sure]} & & \specialcell{i'm {\bf pretty  sure} this is \\ a nordic country}  & \\ \sep 
  
    agreement &  & too, sounds [good] & 
     & ahh {\bf sounds good} to me \\ \sep 

    negations & not$^{\dagger}$ & \specialcell{no$^{\dagger}$, dont } & its definetly {\bf not} that side & \specialcell{i have {\bf no} idea}\\  \midrule

    coordinating conj. & and$^{\dagger}$, but & &  \specialcell{building style {\bf and} cars are also \\ quite uk ish} &  \\ \sep
    subordinating conj. & & though, as$^{\dagger}$ & & \specialcell{{\bf though} it could be a \\forgotten part of Italy} \\ \sep

    demonstrative pron. & these, there$^{\dagger}$, this & that$^{\dagger}$, thats & trust me on {\bf this} one!& lets try {\bf that} one \\ \sep

    articles & the$^{\dagger}$ &  & look at {\bf the} statues  &  \\ 
       filler words &  & ah, haha$^{\dagger}$ & &  no clue {\bf haha}  \\ 
 \midrule
    
    others & \specialcell{it, is$^{\dagger}$, its$^{\dagger}$,  by, from, \\ in$^{\dagger}$, on$^{\dagger}$, somewhere$^{\dagger}$ \\ found, said, saw$^{\dagger}$, \\ far, move } & 
    \specialcell{for, about, have$^{\dagger}$, did$^{\dagger}$, \\ do, lets, looks, makes, \\ put, near, all, your} & &  \\ 
    \bottomrule
  \end{tabular}
    \label{tab:conf_features}
\end{table*}

We formulate this exploration as a paired prediction task: given a pair of equally competent teammates, can we predict from their language which one is more confident?  If such predictive  signals exist, it follows that confidence --- rather than competence alone --- can 
influence
 how people talk when collaborating.

Analyzing the language of the conversations is complicated by the fact that 
utterances
 tend to be short texts that are sprinkled with  misspellings and abbreviations common in online chats; 
see 
Table \ref{tab:chat_example} for an example chat and the last two columns of Table \ref{tab:conf_features} for additional example utterances from individual players.
To find particularly predictive language signals, we consider three types of features and analyze their strength in predicting confidence.

{\bf Word choice (bag of words, or BOW).} 
We
 expect that an individual's choice of words conveys hints as to her confidence 
 level. 
 In order to find predictive word choices that are generalizable to other, non-geography scenarios, 
 we discard words that are tagged as 
nouns  (e.g., ``continent''), proper nouns (e.g., ``Canada''), adjectives (e.g., ``Portuguese'') 
and
a set of manually identified adverbs and verbs.\footnote{
A manual scan suggests that these three word types do indeed cover the majority of geography-related terms, which is our intent, although  helpful domain-independent cues such as the adjective ``sure''
 may unfortunately be excluded  by this procedure as well.} 
 We also discard words that appear in only one puzzle. 
 This filtering process leaves us a vocabulary size of 474 words.

{\bf Ideas.} Figure \ref{fig:per_idea} suggests that an individual's confidence level is reflected in the number of ideas  
she
brings into the discussion, where our notion of ``ideas'' was defined above in Section \ref{sec:interaction:mediate}. Because it may be the case that {\it when} ideas are introduced may also indicate an individual's confidence, 
in addition to (1) the total number of ideas introduced during the entire game, we also 
consider the number of ideas introduced at two special stages: (2) before the start of the discussion, when 
a player
 can provide 
 a
reason for
her initial individual guess (recall that this explanation is made instantly available to  the other teammates once  the team phase commences); and (3) at the very 
early stages of the conversation 
(i.e., within each player's  first three utterances).

{\bf Additional indicators.} We also  examine the use of \emph{hedges} and  expressions of agreement, both of which are,
intuitively, cues for lack of confidence. Hedges, which are expressions of uncertainty or lack of commitment, have been
studied extensively in the linguistics literature (a classic reference is \cite{lakoff1975hedges}), and identifying
hedging is an active area of NLP research \cite{Farkas+al:2010a}. We adopt a list of hedging terms \cite{tan+lee:16}, such as
``apparently'' or ``in my opinion'', 
created by 
by manual curation of prior
collections \cite{Farkas+al:2010a,hanauer2012hedging,hyland1998hedging}. We consider agreement cues because their
occurrence potentially indicates that the speaker was not too confident in their own opinions;
we use the list of agreement cues from \cite{niculae16constructive}.

To set up our competence-controlled paired prediction task, we match 
the most and least confident players in a team (not allowing ties) only if the two players are equally correct in their individual guesses.\footnote{To account for data sparsity and simplify the task we group the two highest confidence levels together. We also considered the task of detecting absolute, rather than relative, confidence and obtained qualitatively similar results.}
This results in 747 matched pairs from 
 216 unique puzzles.
Pair features are simply obtained by taking the difference between the feature values for each of the two players, e.g., for idea features, we take the number of ideas introduced by the first player in the pair subtracted by the number of ideas introduced by the second player in the pair.

We use logistic regression and evaluate performance
 using a leave-one-puzzle-out approach: 
in each fold, we reserve instances from one puzzle as the test
set
and train on instances from all other puzzles; 
this further controls
   for effects of puzzle-specific cues. 
  We report accuracies macro-averaged by puzzle, such that performance on each puzzle has equal weight regardless of how many instances it contains.

With our setup, 
both random guessing and using correctness as a predictor would yield 50\% accuracy. The length of the
utterances, i.e., the number of words a player speaks, is also found not to do better than random guessing. 
BOW features can 
predict significantly better than chance or correctness alone (binomial test $p$-value $<$ 0.001), with a macro-average accuracy of 58\%.
This suggests that there are indeed linguistic signals that are predictive of the difference in confidence between equally-competent teammates. 
We also note that adding the additional 
indicators
 does not bring noticeable improvement. 

Notably, the three ideas features can on their own perform significantly better than chance or correctness ($p$-value $<$ 0.001), achieving 60\% macro-averaged accuracy. 
While all idea introduction features have positive coefficients 
(i.e., introducing more ideas signals more confidence),
 the order of relative importance of these features 
 suggests
  that the early stages of the conversation 
 are
  particularly indicative of relative confidence:

  \begin{center}
  \begin{tabular}{ l  r}
  \toprule
  {\bf feature type} & {\bf coef.} \\ \sep
 number of ideas in reason & 1.05 \\
number of ideas in first 3 utterances & 1.05 \\
number of ideas overall & 0.50 \\
  \bottomrule
  \end{tabular}
\end{center}

 \vskip 1cm

\subsection{Linguistic features of interest}
In order to understand possible
sources of confidence signals beyond the introduction of ideas, 
we now look at the BOW linear regression coefficients.
Table \ref{tab:conf_features} shows words that ranked among the top 30 features in at least half of the folds, grouped into post-hoc categories to emphasize several distinctions between 
the language used by more-confident vs. less-confident individuals.

In terms of the use of personal pronouns (first row of Table \ref{tab:conf_features}), 
more confident individuals are more likely to speak on behalf of the team, whereas 
less confident individuals tend to stick more with first and second person perspectives. 
Interestingly, this echoes in part findings showing that in small groups leaders are generally more other-focused, preferring first-person plural to first-person singular \cite{cassell2006language, kacewicz2014pronoun}.
In this regard, more confident individuals seem to assume an ad-hoc leadership role (even when they are equally-competent).

On the other hand, less confident players tend to consult others' opinions more, as indicated by the increased use of interrogatives (second row). We also note that ``did'' and ``do'' categorized under ``others'' are mostly used in questions involving such interrogatives.

When expressing thoughts (third row), 
less confident individuals tend to use past tense,\footnote{
In these chats, past continuous tense (``was thinking'') is more prevalent than present continuous tense (``am thinking'').
} which might be an indication that they perceive themselves as less authoritative members in the dialogue, according to some studies in linguistics \cite
{Johnstone:Linguistics:1987}. 

The patterns for features that are intuitively, or at least anecdotally, associated with confidence, such as hedging, certainty and  agreement, are listed in the second block of rows in Table \ref{tab:conf_features}.
These patterns do not align completely with common intuition. 

More confident individuals do indeed tend to use more phrases indicating certainty, and 
express less agreement with other's opinions. On the other hand, hedges and modal verbs, are not necessarily reserved to the less confident people. For instance, ``might'' and ``probably'' are 
associated with the more confident people; one hypothesis 
worth exploring further is that
   confident individuals make use of certain types of hedges as a strategy to gently persuade their partners~\cite{tan+etal:16a}. 

There are also interesting differences in terms of the surface structure of the utterances (third block) with 
more confident individuals being more likely to use coordinating conjunctions and the definite article ``the'', potentially in an attempt to bring more concrete
evidence into the discussion.  On the other hand, the utterances of 
less confident individuals are sprinkled with filler words.

This shallow post-hoc analysis of words signaling confidence (or lack thereof) can serve as a starting point for future work on more complex linguistic and conversational features for confidence detection.  The fact that such predictive cues exist even in a setting that controls for competence underlines the role confidence has in shaping people's language in group discussions.  Furthermore, this suggests that future interfaces for facilitating group decision-making could potentially extract confidence information from the language of the participants.

\vskip 0.5 in

\section{Additional Related Work}
\label{sec:related}
Confidence has been studied as a factor affecting human behavior. 
Horowitz \cite{Horowitz:JPsychol:1966} has shown that there is a significant negative correlation between a person's level of imitation behavior and the difference between the person's confidence relative to others.  
Confidence as approximated by poll scores is shown to correlate with how frequent candidates shift topics in political debates \cite{Prabhakaran:AclJointWorkshopOnSocialDynamicsAnd:2014}. In the group opinion forming processes, 
highly confident individuals 
have been identified as one of
the two major attractors of opinion \cite{Moussaid:PlosOne:2013}, based on simulations on models derived
from controlled experiments. 
However, most of these these experiments merely present participants the opinions and confidence levels of others, 
without allowing any real interactions between them. 

The correlation between confidence and \correctness has been studied in 
many contexts. 
Studies relating
the confidence of eyewitness to the actual \correctness
\cite{bothwell1987correlation} and
self-reported confidence to observed competence of junior medical officers \cite{barnsley2004clinical}
both indicate important problems with the use of confidence as a predictor of correctness. Miscalibration has been noted for a long time, with most studies focusing on overconfidence \cite{alicke1985global, Dunning01062003, kruger1999unskilled}. See Moore and Healy \cite{moore2008trouble} for a reconciliation of existing studies on overconfidence.

\section{Conclusions and Future Work}
\label{sec:discussion}

Our main contribution in this paper is to investigate pre-existing and new research questions regarding the effects of confidence, particularly when misaligned with correctness, on group decision-making and group discussions, doing so at a large scale and in a natural online setting.

Our results,
as well as the limitations of our setting,
 point naturally to
several directions for future work:

{\bf Confidence prediction and miscalibration detection.} In our explorations of  possible linguistic cues of confidence, we simplified matters by 
binarizing confidence labels (more confident vs. less confident) and
applying basic cue-extraction techniques. 
Given that we 
do
discover that there are signals in language that capture confidence, one natural extension is to aim at more accurately predicting confidence at a finer-grained level, perhaps by applying more sophisticated natural language processing techniques.
We could build on such work to address the more ambitious goal of detecting misalignment between confidence and correctness, with the eventual aim of facilitating group discussion by ameliorating the effects of 
under- and over-confidence 
on team conversations and outcomes.

{\bf Finer-grained measures of confidence and competence. } In this work, we discretized players' confidence and competence using four coarse levels.  As discussed in Section 3, this discretization allows us to identify clear cases of under- and over-confidence. In future work, finer-grained measures of confidence and competence might enable better controls, while providing a more detailed analysis of the extent of confidence-competence (mis)alignment effects in team interactions. 

{\bf Dynamics of team discussions.} We have looked into conversational dynamics in terms of idea introduction and selection, where we employed a simple heuristic to track ideas.
Pursuing better methods of representing ideas, capturing idea introduction, identifying idea adoption, and further distinguishing good and bad ideas would be beneficial in further understanding the dynamics of discussions. 

On a related note, in this work we
have only considered the initial confidence of individual players. 
Studies have looked into how individual confidence combines to form eventual dyadic confidence in collaborative decisions \cite{schuldt2015confidence}, and it would be interesting to track and understand how individuals' confidence in their own guess and in their team's working solution changes over time. 

{\bf Causality. } We have revealed  correlations between confidence and influence in teams, as well as between confidence and team synergy arising from interaction. Our full control over the game interface gives us the ability to  conduct experiments involving the introduction of stimuli that artificially manipulate individuals' confidence levels; this ability opens the possibility of studying not just correlation, but also causality.

{\bf Confidence in other decision-making contexts.} Will our findings generalize beyond StreetCrowd? While we have striven to analyze our data in a non-domain specific fashion, studying the effects of confidence in team decision-making in other contexts,
such as group work in Massive Open Online Courses, or in the process of eRulemaking, is an important direction for future work. 

\vskip 0.1 in

\xhdr{Acknowledgments} 
We are grateful to 
Eric Horvitz and 
Sendhil Mullainathan for inspiring discussions on the role of confidence in conversations. 
We would also
like to thank
Pantelis Analytis,
David Byrne,
Claire Cardie,
Susan Fussell,
Mala Gaonkar,
Malte Jung,
Dan Jurafsky,
Jon Kleinberg,
Anita W. Woolley,
Justine Zhang, 
and the anonymous reviewers (all of whom were supremely competent!) for their insightful (and confident) comments, and Chenhao Tan for sharing the list of hedges.  
We are also grateful to all StreetCrowd players and to
David Garay, 
Maheer Iqbal, 
Jinjing Liang, 
Vlad Niculae, and
Neil Parker
for participating in the game's development together with the first author.  This work was supported in part by a Discovery and Innovation Research Seed Award from the Office of the Vice Provost for Research at Cornell.

\vskip 0.25 in 

\bibliographystyle{abbrv}
\bibliography{refs}
\end{document}